\documentclass[times, final, 10pt, twocolumn]{elsarticle}
\usepackage{amssymb}
\usepackage{booktabs}
\usepackage{amsmath}
\usepackage{multirow}
\usepackage{xcolor}
\usepackage{makecell}
\usepackage{hyperref}
\journal{Pattern Recognition}

\begin{document}

\begin{frontmatter}

\title{Zero-Shot Skeleton-based Action Recognition with Dual Visual-Text Alignment}

\author[add1,add2]{Jidong~Kuang}
\author[add3,add4]{Hongsong~Wang\corref{cor}}
\author[add1,add2]{Chaolei~Han}
\author[add5,add6,add7]{Yang~Zhang}
\author[add1,add2,add8]{Jie~Gui\corref{cor}}  

\address[add1]{School of Cyber Science and Engineering, Southeast University, Nanjing, China}
\address[add2]{Engineering Research Center of Blockchain Application, Supervision And Management (Southeast University), Ministry of Education, China}
\address[add3]{School of Computer Science and Engineering, Southeast University, Nanjing, China}
\address[add4]{Key Laboratory of New Generation Artificial Intelligence Technology and Its Interdisciplinary Applications (Southeast University), Ministry of Education, China}
\address[add5]{School of Computer Science and Software Engineering, Shenzhen University, Shenzhen 518060, China}
\address[add6]{National Engineering Laboratory for Big Data System Computing Technology, Shenzhen University, Shenzhen 518060, China}
\address[add7]{Guangdong Key Laboratory of Intelligent Information Processing, Shenzhen University, Shenzhen 518060, China}
\address[add8]{Purple Mountain Laboratories, Nanjing 210000, China}

\cortext[cor]{Corresponding author}

\begin{abstract}
Zero-shot action recognition, which addresses the issue of scalability and generalization in action recognition and allows the models to adapt to new and unseen actions dynamically, is an important research topic in computer vision communities. The key to zero-shot action recognition lies in aligning visual features with semantic vectors representing action categories. Most existing methods either directly project visual features onto the semantic space of text category or learn a shared embedding space between the two modalities. However, a direct projection cannot accurately align the two modalities, and learning robust and discriminative embedding space between visual and text representations is often difficult. To address these issues, we introduce Dual Visual-Text Alignment (DVTA) for skeleton-based zero-shot action recognition. The DVTA consists of two alignment modules—Direct Alignment (DA) and Augmented Alignment (AA)—along with a designed Semantic Description Enhancement (SDE). The DA module maps the skeleton features to the semantic space through a specially designed visual projector, followed by the SDE, which is based on cross-attention to enhance the connection between skeleton and text, thereby reducing the gap between modalities. The AA module further strengthens the learning of the embedding space by utilizing deep metric learning to learn the similarity between skeleton and text. Our approach achieves state-of-the-art performances on several popular zero-shot skeleton-based action recognition benchmarks. \textcolor{black}{ The code is available at:} \href{https://github.com/jidongkuang/DVTA}{https://github.com/jidongkuang/DVTA}.

\end{abstract}


\end{frontmatter}

\section{Introduction}
Action recognition emerges as a hot topic in research due to its wide applications in areas such as video understanding, human-computer interaction, health monitoring and robotics. However, with the rapid increase in the number of videos generated due to the prevalence of social media, smartphones, and surveillance cameras, collecting video action labels becomes increasingly impractical and burdensome. Consequently, Zero-Shot Action Recognition (ZSAR) \cite{xu2017transductive} emerges as a viable solution. ZSAR alleviates the burden of collecting labeled data by leveraging a shared semantic space to generalize knowledge across unseen action categories, thus enabling recognition of actions without requiring explicit training examples. 

The key to implementation of ZSAR lies in effectively utilizing the shared semantic space to achieve recognition of unseen classes. A common approach is the projection-based method. For instance, the early method \cite{xu2015semantic} projects low-level visual features into a semantic space composed of word embeddings. However, these methods solely relying on direct visual projection struggle to bridge the gap between modalities. Another typical way is to generate an intermediate space for visual-semantic fusion. For example, BiDiLEL \cite{wang2017zero} employs two-stage mapping, separately mapping visual and semantic features into a latent space. These approaches face the challenge of designing an effective embedding space that can capture the nuances and complexities of visual and text modalities. Moreover, the above methods utilize RGB video data, which is susceptible to background interference, leading to suboptimal performance in ZSAR.

\begin{figure}
    \small
    \centering
    \includegraphics[width=1\linewidth]{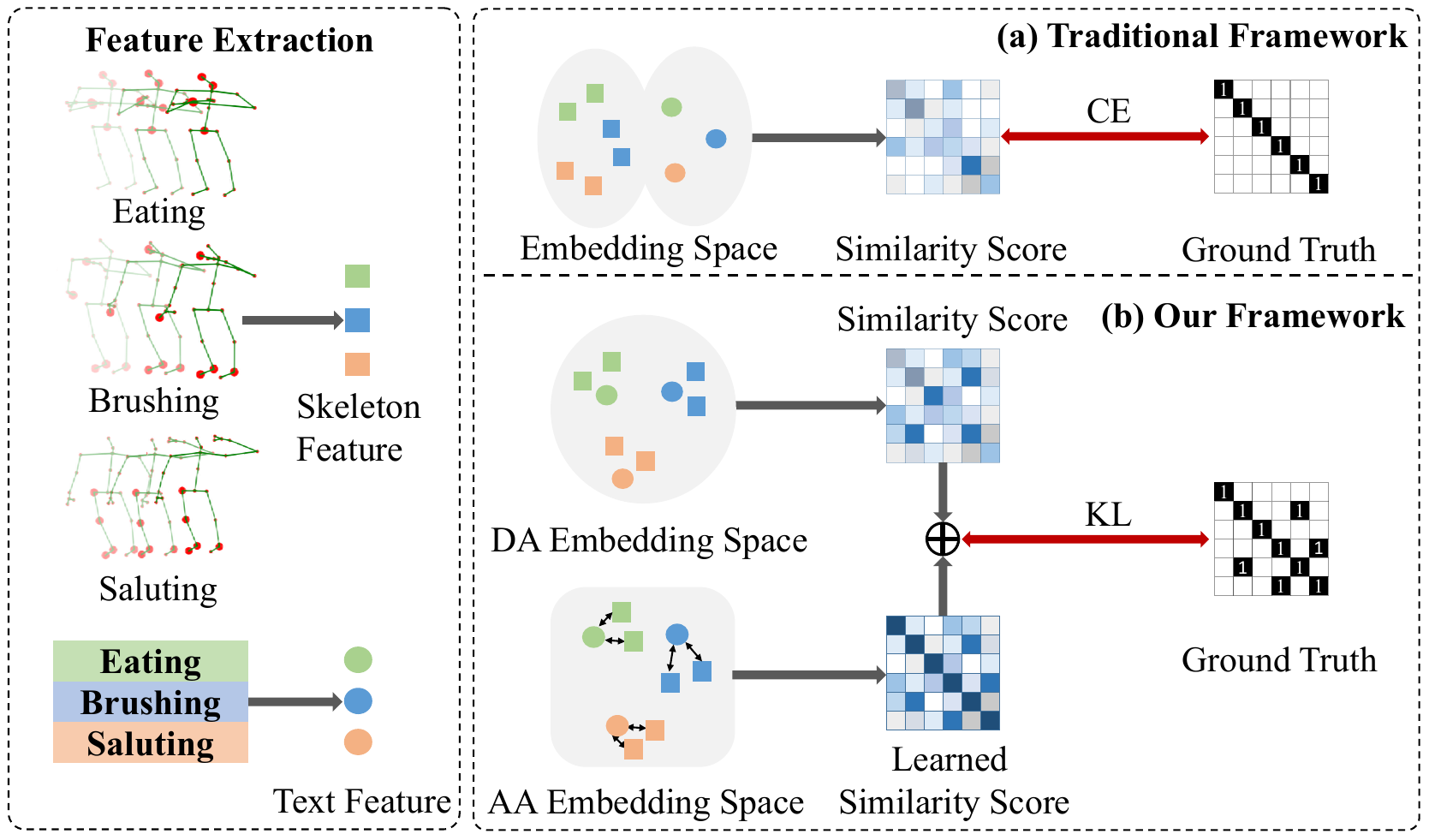}
    \caption{(a) Traditional skeleton-based zero-shot action recognition methods use a single alignment model to align the two modalities.
(b) Our DVTA approach employs a dual alignment strategy, where Direct Alignment is used for initial alignment to strengthen the association between modalities, and distribution alignment is further applied to enhance generalization to unseen classes. KL divergence is employed to generate more positive examples, facilitating the joint optimization of the two spaces.}
    \label{fig:intro_fig}
\end{figure}

With the emergence of depth sensors and advancements in pose estimation algorithms, skeleton-based action recognition \cite{wang2017modeling,weng2025usdrl} has gained attention. Skeleton data is robust to lighting and environmental changes, making it a good alternative to RGB video data. However, compared to RGB video data, there is relatively less research on ZSAR using skeleton data. Recent methods perform distribution alignment to enhance modality correspondence. To capture complex correlations between the distributions of the visual space and the text semantic space, SMIE \cite{zhou2023zero} addresses skeleton-based ZSAR via mutual information estimation and maximization. However, these methods often overlook the significant differences between the two spaces and directly use the connection models on the raw extracted features to capture dependencies, which limits their performance to a certain extent.

To effectively reduce the discrepancy between visual and semantic modalities, we refrain from directly applying connection models to learn the similarity between the two spaces. Instead, we introduce a dual alignment strategy, as depicted in Figure~\ref{fig:intro_fig}. The initial step involves mapping features from distinct modalities into a shared embedding space, followed by executing metric learning within that common space. 

Specifically, we propose the Dual Visual-Text Alignment (DVTA) network, which employs two key components for modality alignment. The first component is Direct Alignment (DA), which utilizes a specially designed deep neural network for visual projection to map skeleton features into the semantic space. An attention-based Semantic Description Enhancement (SDE) module is then used to integrate visual information into text features, which establishes a closer connection between skeleton sequences and text descriptions. During training, a supervised training loss is formulated to generate more positive examples, maximizing the similarity between the paired skeleton and text features within this space. The second component is the Augmented Alignment (AA) module, which uses a neural network to further align distributions by learning similarity scores between different modality features within the latent space. The LeakySigmoid activation function is designed to normalize the predicted similarity values. The KL divergence loss is employed to jointly optimize the alignment modules, enhancing the alignment between the two modalities.

Our main contributions can be summarized as follows: 
\begin{itemize}
    \item We propose a Dual Visual-Text Alignment (DVTA), a novel zero-shot approach for skeleton-based action recognition. The method enhances generalization to unseen classes by jointly optimizing two modules: Direct Alignment (DA) and Augmented Alignment (AA).
    \item The proposed DA module facilitates cross-modal understanding and alignment through a deep visual projector and the cross-attention-based Semantic Description Enhancement (SDE). The AA module further aligns distributions by leveraging the deep metric learning with the LeakySigmoid function.
    \item Extensive experiments demonstrate the superior performance of the proposed method on the NTU RGB+D, NTU RGB+D 120, and PKU-MMD datasets.
\end{itemize}

\section{Related Work}
\subsection{Zero-Shot Action Recognition}
With the rapid development of large-scale language models, new categories can be recognized using a shared semantic space, leading to an increasing interest in zero-shot action recognition. Currently, most approaches are dedicated to aligning the visual space \cite{han2020learning} and the semantic space \cite{brattoli2020rethinking}. Numerous studies \cite{gao2023learning,brattoli2020rethinking} construct an end-to-end framework to achieve the transformation from the original visual input to the final visual semantic representation in a unified way. In order to capture inter-modal correlations more accurately, attentional mechanisms are designed to focus on learning the multimodal space \cite{qi2023energy}. In addition, augmented category information methods \cite{lin2023match} can further improve the performance of zero-shot learning. Recently, DeCalGAN \cite{wang2023deconfounding} generates unseen action video features to transfer knowledge from known action categories to the feature distribution of unknown categories. 

\subsection{Skeleton-Based Action Recognition}
We primarily review methods based on Graph Convolutional Network (GCN), which are used to extract skeleton features. Yan et al.~\cite{yan2018spatial} propose the innovative ST-GCN, which represents human joints as graph structures and applies spatio-temporal GCN for the first time to skillfully capture the spatial and temporal features of the human skeleton. This pioneering work leads to the thriving development of GCN-based skeleton action recognition methods~\cite{dai2023global,zhao2024sharing,yin2024spatiotemporal}.
Li et al. \cite{li2019actional} introduce the actional-structural graph convolution network (AS-GCN), which generates a generalized skeleton graph to capture richer dependencies between joints.
Shi et al.~\cite{shi2019two} propose 2s-AGCN, which utilizes second-order information of skeleton data to enhance recognition accuracy and dynamically learns graph topology to improve the model's generality. To further improve efficiency, Cheng et al. \cite{cheng2020skeleton,cheng2021extremely} propose the shift graph convolutional network (Shift-GCN) to significantly reduce the computational complexity of the GCN method. Subsequently, they also introduce a more lightweight Shift-GCN++ \cite{cheng2021extremely} to further optimize performance. Chen et al. \cite{chen2021channel} develop a Channel-wise Topology Refinement Graph Convolution (CTR-GCN) to dynamically learn the topology and capture the inter-node relationships in a more fine-grained manner. Chi et al. \cite{chi2022infogcn} adopt an information bottleneck to guide the model in learning latent representations, and introduce attention-based graph convolution to capture contextual intrinsic topology. Recently, Xiang et al. \cite{xiang2023generative} propose a Generative Action-description Prompts (GAP) approach, which utilizes action semantic information to enhance the representation learning ability of the skeleton encoder. \cite{qiu2024multi} and~\cite{wu2023spatiotemporal} use the multi-granularity information of skeleton to improve recognition ability. These research works achieve satisfactory performance in skeleton feature extraction, providing strong support for encoders in the field of skeleton-based action recognition.
\subsection{Zero-Shot Skeleton-Based Action Recognition}
Aligning the skeleton space with the semantic space without accessing unseen class samples for training poses a significant challenge, resulting in relatively few works on zero-shot skeleton action recognition. Zero-shot learning methods previously applied to other domains are creatively extended to this domain. For example, DeViSE \cite{frome2013devise}, originally designed for visual image classification tasks, maps extracted skeleton features to the text feature space through a learnable linear projection. ReViSE \cite{hubert2017learning} employs the maximum average discrepancy loss function to further facilitate the alignment between the two modalities. RelationNet \cite{jasani2019skeleton} calculates the similarity between the projected features using the attribute network and the relation network. Subsequently, JPoSE \cite{wray2019fine} proposes the construction of a separate multimodal space for each PoS tag to complete cross-modal fine-grained action retrieval. CADA-VAE \cite{schonfeld2019generalized} utilizes a variational autoencoder to learn the latent space between modalities. More recently, works specifically targeting zero-shot skeleton-based action recognition have emerged. For instance, SynSE~\cite{gupta2021syntactically} leverages syntactic structure partition labels to align skeleton features with corresponding verb embeddings and noun embeddings. SMIE~\cite{zhou2023zero} aligns the distributions of the two modalities by maximizing mutual information. Some works utilize large-scale language models to generate extensive semantic information for modality alignment. For instance, PURLS~\cite{zhu2024part} and STAR~\cite{chen2024fine} employ large-scale language models to create multi-part descriptions for fine-grained alignment, while InfoCPL~\cite{xu2025information} combines numerous descriptive texts to enrich semantic information, improving recognition performance. However, these approaches rely heavily on the manual preparation of extensive text information, which is impractical for real-world applications.
Our approach differs from these methods in two key ways. First, we focus on structural innovation by designing a specialized deep visual projector to bridge the substantial gap between skeleton and text features. Additionally, we propose a dual alignment strategy that jointly optimizes two alignment networks to strengthen the correlation between skeleton and text features. Second, we enrich semantic information using only a minimal set of label descriptions, significantly reducing the need for manual preparation.

\section{Method}
Zero-shot action recognition aims to recognize unseen action categories in the test set, despite the absence of corresponding training skeleton samples. The key to achieving this lies in aligning the visual feature space with the text semantic space, leveraging the joint embedding space to recognize the action of samples from unknown categories. Nevertheless, aligning visual and text remains a challenging task due to the semantic gap between the skeleton sequence and the corresponding text category. To bridge this gap, we introduce a Dual Visual-Text Alignment (DVTA), which mainly comprises two alignment modules, Direct Alignment (DA) and Augmented Alignment (AA).

\begin{figure*}[t]
    \centering
    \includegraphics[width=0.9\linewidth]{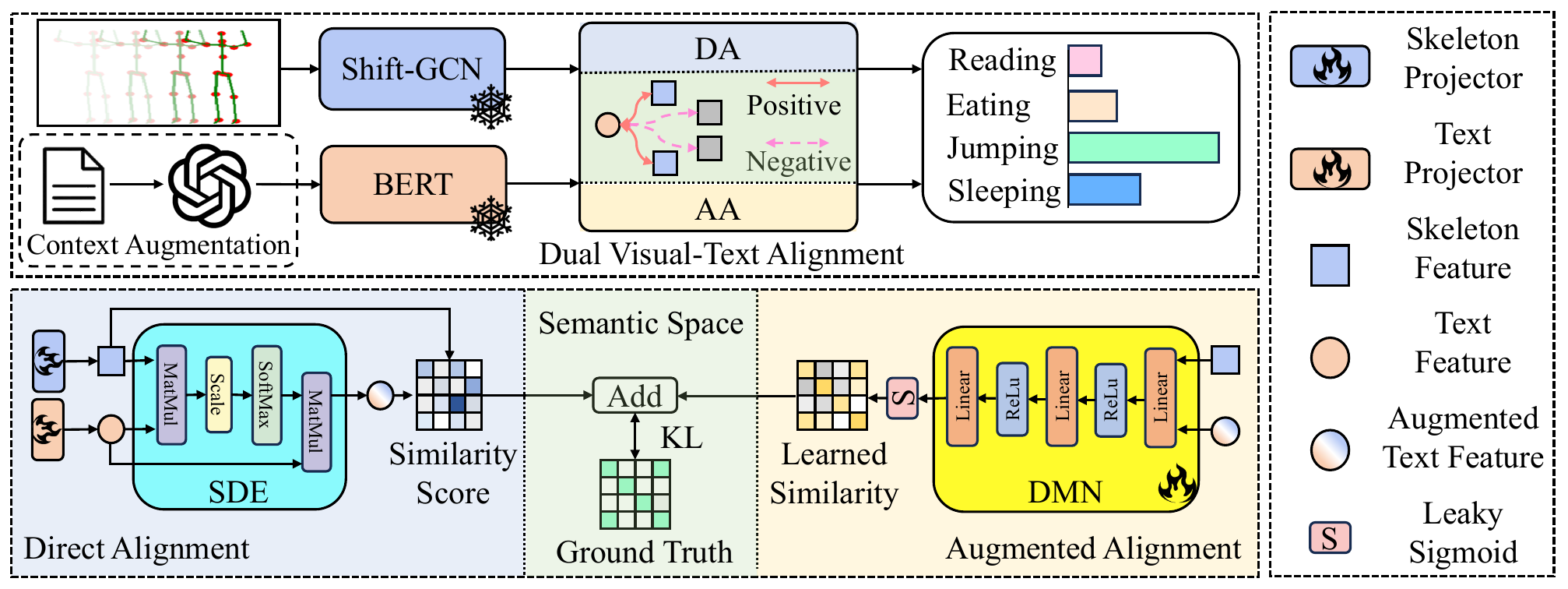}
        \caption{\textcolor{black}{Framework of the proposed Dual Visual-Text Alignment (DVTA). It jointly optimizes two alignment modules—Direct Alignment (DA) and Augmented Alignment (AA)—in a supervised manner.}}
    \label{fig:framework}
\end{figure*}

\textcolor{black}{Figure \ref{fig:framework} illustrates the overall structure of the proposed method. Specifically, the DA module maps skeleton features and text features to a common latent space using different projection modules. Subsequently, the Semantic Description Enhancement (SDE) submodule, which utilizes the cross-attention mechanism, generates augmented text features containing visual information, thereby fostering a closer connection between the skeleton sequence and the text description. The AA module computes the correlation between the two modalities using a designed Deep Metric Network (DMN) and further aligns their distributions by normalizing the similarity scores through the LeakySigmoid activation function. The KL divergence loss incorporating a greater number of positive examples is employed as the contrastive loss to jointly optimize the two alignment modules, thereby enhancing the alignment between the two modalities.}
\subsection{Visual-Text Embedding} \label{sec:JEL}
\begin{figure}[t]
    \centering
    \includegraphics[width=1\linewidth]{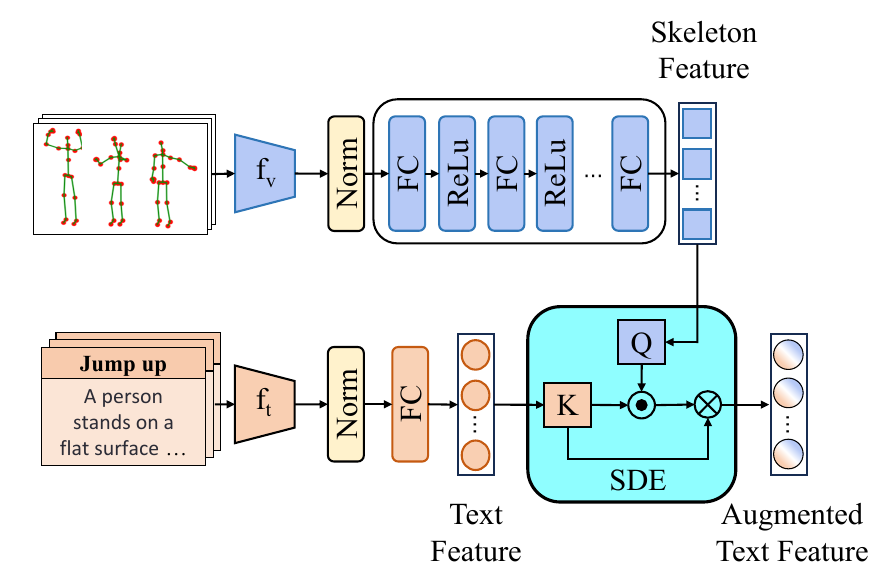}
    \caption{The process of Visual-Text Embedding. It involves using a designed deep visual projector to map skeleton features into a shared semantic space, followed by obtaining augmented text features through the SDE sub-module.}
    \label{fig:JEL}
\end{figure}

Previous methods for cross-modal projection often rely on projecting visual features into the semantic space using a single linear layer. However, due to the significant differences between skeleton and text features, this simple projection is ineffective. To address this issue, we propose an approach that utilizes a specially designed deep neural network to map skeleton features to the semantic space, thereby more effectively learning and reducing the gap between the two modalities. Simultaneously, a linear layer is retained for processing text features to preserve the richness of the text descriptions. The entire Visual-Text Embedding process is illustrated in Figure~\ref{fig:JEL}. After passing the embedded vectors through the designed projector, the SDE submodule is applied to obtain augmented text features.

\subsubsection{Deep Feature Projection}
Let $({x^s},{y^s})$ be a sample from \textbf{seen classes} in the training dataset, where ${y^s}$ represents action labels of seen classes and ${x^s}$ represents the corresponding skeleton sequence. Similarly, the sample from \textbf{unseen classes} in the testing dataset is denoted as $({x^u},{y^u})$. Using visual feature extractor ${f_v}$ and text feature extractor ${f_t}$, visual feature $v$ and text feature $t$ are obtained as:
\begin{equation}
    {v^i} = \mathrm{norm}({f_v}({x^i})), \; {t^i} = \mathrm{norm}({f_t}({y^i})), \; i \in \{ s,u\},
\end{equation}
where $\mathrm{norm}$ denotes the L2 normalization. Feature normalization is used before feeding the features to projection networks to enhance stability.
The visual feature extractor ${f_v}$ is pre-trained on seen class samples, while the text feature extractor ${f_t}$ is pre-trained on a large-scale language corpus. 

The features $v$ and $t$ come from different spaces and the dimensions are also different. Two projection networks are designed to map the features of skeleton and text to a common embedding space. In order to map the skeleton features to the same dimension as the text features to a greater extent, a learnable depth nonlinear mapping projector is used to transform the skeleton features to the common embedding space, the effect of which will be reflected in section~\ref{sec:ablation}. A linear layer is utilized to map text features to the common space. 
Let $F_v$ and $F_t$ be the projectors for skeleton features and text features, respectively. To simplify the expressions, the superscripts for visible and unseen classes are omitted. The transformed skeleton and text features are:
\begin{equation}
    v_e = F_s(v), \; t_e = F_t(t),
\end{equation}
where the feature dimensions of both $v_e$ and $t_e$ are $h$, and $h$ refers to the embedding dimension size.
\subsubsection{Semantic Description Enhancement}
Through the cross-attention mechanism, the SDE module allows the model to create a tighter connection between skeleton and text features, thus better capturing the relationship between the two modalities.

In zero-shot action recognition, the accuracy and richness of semantic space information are critical for the generalization to unseen samples. The inspiration for semantic description enhancement comes from the complex process in the human brain, where both action relevance and environmental context are considered during action recognition. By enriching the semantic content of actions through context augmentation, the model's recognition capability can be improved.
We use GPT-3~\cite{brown2020language} to generate contextual descriptions for action labels by employing a fixed prompt template: ``Describe the context in which the $<$action$>$ occurs''. For example, for the action ``eat meal'', a descriptive sentence might be: ``The person is sitting at a table, facing a plate of food in front of them.''. This context description, after processing through the above text feature extraction operations, yields the action context text feature $t_{cont}$.
We base on a cross-attention mechanism to jointly embed the visual feature \( v_e \), label text feature \( t_e \), and context text feature \( t_{cont} \) to generate the semantically augmented text feature \( t_{aug} \). Specifically, we define \( K \) and \( V \) as \( T = stack({t_e}, {t_{cont}}) \in {\mathbb{R}^{2 \times h}} \), and \( Q = v \) as the visual query. The semantically augmented text feature \( t_{aug} \) can be obtained as:
\begin{equation}
{t_{aug}} = softmax(\frac{{Q{K^T}}}{{\sqrt h }})V.
\end{equation}
The augmented text feature \( t_{aug} \) is used for subsequent contrastive learning.

\subsection{Dual Alignment Network}
After the Visual-Text Embedding, we obtain augmented text features enriched with visual information, facilitating a closer connection between the visual and text modalities. Subsequently, we apply a dual alignment strategy to these features. First, Direct Alignment is employed to reduce the gap between different modalities,  followed by the Augmented Alignment module to achieve a deeper level of distribution alignment. Throughout the training process, the KL divergence loss is used to jointly optimize these two alignment modules.
\subsubsection{Direct Alignment}
Given that the number of skeleton feature samples significantly exceeds the number of action labels, it is reasonable to expect that a batch of data will contain a large number of skeleton samples corresponding to a single label, indicating the presence of multiple positive examples. Therefore, the cross-entropy loss is inappropriate. Instead, we utilize KL divergence as the skeleton-text contrastive loss, allowing each skeleton feature sample to form multiple positive examples with the corresponding same-label text features, as referenced in~\cite{wang2021actionclip,xiang2023generative}. The similarity score between the skeleton features and text features can be calculated as follows:

\begin{equation}
\begin{array}{l}
p_{1}^{v2t}({v_i}) = \frac{{\exp (sim({v_i},{t_i})/\tau )}}{{\sum {_{j = 1}^B\exp (sim(v_j,t_j)/\tau )} }},\\
p_{1}^{t2v}({t_i}) = \frac{{\exp (sim({t_i},{v_i})/\tau )}}{{\sum {_{j = 1}^B\exp (sim(t_j,v_j)/\tau )} }},
\end{array}
\label{eq:simlarity1}
\end{equation}
where \( (v_i, t_i) \) represents a positive example formed by the visual feature and its corresponding text feature, while \( (v_i, t_j) \) forms a negative example pair when \( j \neq i \). $sim({v_i},{t_i})$ represents cosine similarity, $B$ denotes batch size, $\tau$ is a learnable temperature parameter, and $p_{1}^{v2t}({v_i})$ and $p_{1}^{t2v}({t_i})$ denote the skeleton-to-text and text-to-skeleton similarities, respectively.

The Direct Alignment module alone can be used for zero-shot skeleton-based action recognition. During training, the objective loss $L_1$ is:
\begin{equation}
    {L_1} = {{\bf{{\rm E}}}_{{v},{t} \sim D}}\left[ {KL({p_{1}^{v2t}}(v),{y^{v2t}}) + KL({p_{1}^{t2v}}(t),{y^{t2v}})} \right],
\end{equation}
where $D$ represents the training dataset, ${y^{v2t}}$ and ${y^{t2v}}$ respectively denote the true similarity scores, where the probability of positives is 1 and the probability of negatives is 0. By contrasting the skeleton-text pairs within batches, we jointly optimize the two projection networks.

\subsubsection{Augmented Alignment}
We first concatenate the projected skeleton feature $v$ and text feature $t$, then feed the concatenated feature to the Deep Metric Network $E$ to predict similarity score $G({v},{t})$ between the two features:
\begin{equation}
    G({v},{t}) = LeakySigmoid(E({v},{t})),
\end{equation}
where the LeakySigmoid is an activation function which transforms the value output by $E$ to the range of similarity scores of $[0,1]$. The LeakySigmoid activation function combines the smoothness of the Sigmoid function with the nonzero gradient property of LeakyReLU. The formula of LeakySigmoid is defined as:
\begin{equation}  
\text{LeakySigmoid}(x) =   
\begin{cases}   
\frac{1}{1 + \exp(-x)}, & \text{if } x > 0 \\  
\gamma \cdot \exp(\gamma \cdot x ), & \text{if } x \leq 0   
\end{cases}  
\end{equation}
where $\gamma$ is a hyperparameter that controls the slope of the function when the input is negative. This parameter incorporates a ``leak'' into the Sigmoid function, ensuring a nonzero gradient even for negative input values. Consequently, it effectively mitigates the vanishing gradient issue commonly encountered during the training of deep neural networks. 
In practice, the value of $\gamma$ is typically set to a very small positive number (e.g., 0.001 or 0.01) to still maintain some gradient when the input is negative, but not significantly affect the behavior of the function when the input is positive. 

Similar to Eq.~\ref{eq:simlarity1}, the similarity scores between the skeleton sequence and the text can be obtained:
\begin{equation}
\begin{array}{l}
p_2^{v2t}({v_i}) = \frac{{\exp (G({v_i},{t_i})/\tau )}}{{\sum {_{j = 1}^B\exp (G(v_j,t_j)/\tau )} }},\\
p_2^{t2v}({t_i}) = \frac{{\exp (G({t_i},{v_i})/\tau )}}{{\sum {_{j = 1}^B\exp (G(t_j,v_j)/\tau )} }}.
\end{array}
\label{eq:simlarity2}
\end{equation}

The two similarity scores obtained in Eq.~\ref{eq:simlarity1} and Eq.~\ref{eq:simlarity2} can be combined to obtain the final similarity score:
\begin{equation}
{p^{v2t}} = (p_1^{v2t} + p_2^{v2t})/2,{p^{t2v}} = (p_1^{t2v} + p_2^{t2v})/2.
\end{equation}

Trained by KL divergence contrastive loss, the network $E$ is able to make the skeleton features of positive pairs have a higher similarity with the text features, while reducing the similarity of negative pairs. This process effectively aligns the distributions of the two features, enabling the model to generalize better to unseen classes for zero-shot learning. The final loss is obtained as follows:
\begin{equation}
L = {{\bf{{\rm E}}}_{{v},{t} \sim D}}\left[ {KL({p^{v2t}}(v),{y^{v2t}}) + KL({p^{t2v}}(t),{y^{t2v}})} \right].
\end{equation}

During testing, ${p^{s2t}}$ is computed by comparing the skeleton sequence with the text features of each unseen class, so as to match the most similar class.

\section{Experiments}
\subsection{Datasets}

\noindent\textbf{NTU RGB+D 60}: The NTU RGB+D 60~\cite{shahroudy2016ntu} dataset is captured by three Microsoft Kinect cameras at three different angles and is often used as a skeleton-based human action recognition dataset. It contains 60 different action categories, each performed by 1 or 2 people. There are a total of 56,880 samples, each containing 25 skeleton joint point 3D coordinates. The standard splits for the dataset consist of two types: 1) Cross-Subject (X-Sub): the dataset is divided according to the ID of the volunteers, half of the volunteers form the training set and the rest form the test set. 2) Cross-View (X-View): the samples captured by camera views 2 and 3 form the training set, and those captured by view 1 form the test set.

\noindent\textbf{NTU RGB+D 120}: The NTU RGB+D 120~\cite{liu2019ntu} dataset extends the action classes of NTU RGB+D 60 to 120 classes, adding 57,367 skeleton sequence samples. Two official evaluation benchmarks are provided: 1) Cross-Subject (X-Sub): the test set and training set are divided according to different subjects. 2) Cross-Setup (X-Setup): the dataset is divided by different camera setups.

\noindent\textbf{PKU-MMD}: The PKU-MMD~\cite{liu2017pku} dataset consists of 51 distinct action categories, comprising nearly 20,000 skeleton sequence samples. The dataset provides two official partition settings: (1) Cross-Subject (X-sub): The training and testing sets are divided based on different subjects. (2) Cross-View (X-view): The dataset is partitioned according to varying views captured by Kinect devices.
\subsection{Implementation Details}

For data preprocessing, we resize the skeleton sequence to 50 frames using linear interpolation and removal of invalid frames, which is the same as on Cross-CLR \cite{li20213d}. For text modality, 768-dimensional word embeddings are obtained by a pre-trained text feature extractor from Sentence-Bert. For skeleton modality, Shift-GCN \cite{cheng2020skeleton} are used as feature encoders to obtain 256-dimensional features. The dimensionality is then increased to 768 dimensions by the nonlinear feature projector. The semantic features are passed through a linear layer to enhance the learning of similarity and to retain as much semantic information as possible. The optimizer employed is Adam, with a learning rate of 1e-5, and the training process spans 100 epochs, utilizing the CosineAnnealing scheduler. The mini-batch size is 128, and the temperature parameter for the contrastive loss is set to 0.1. Our implementation relies on the PyTorch framework, and all experiments are carried out using an NVIDIA GeForce RTX 4090.

Our approach is compared with recent state-of-the-art approaches of skeleton-based zero-shot learning, e.g., SMIE \cite{zhou2023zero}, SynSE \cite{gupta2021syntactically}, CADA-VAE \cite{schonfeld2019generalized} and JPoSE \cite{wray2019fine}. SMIE \cite{zhou2023zero} and SynSE \cite{gupta2021syntactically} strive to use a connection model to associate two modalities, while CADA-VAE \cite{schonfeld2019generalized} and JPoSE \cite{wray2019fine} aim to learn a common latent space.


\subsection{Experimental Results}

\noindent\textbf{Comparison with State-of-the-arts}: We compare the proposed DVTA with the state-of-the-art approaches to demonstrate its effectiveness in zero-shot learning. To ensure a fair comparison, our experimental setup strictly remains consistent with that of SynSE~\cite{hubert2017learning}, including the use of identical skeleton feature and semantic feature encoders as well as the same split between seen and unseen classes. This setting incorporates only two datasets: NTU-60 and NTU-120. We employ Shift-GCN~\cite{cheng2020skeleton} as the skeleton feature extractor and leverage the pre-trained Sentence-Bert for semantic feature extraction. Follwing SynSE~\cite{hubert2017learning}, we establish two distinct category divisions for each dataset. The NTU-60 dataset divided its 60 categories into 5 and 12 unseen classes, corresponding to the 55/5 and 48/12 splits in the table. For the NTU-120 dataset, which boasts a larger sample size and more categories, the number of unseen classes increased to 10 and 24, resulting in the 110/10 and 96/24 splits. 
Table~\ref{tab:sota} presents a comparison of the performance of the DVTA with state-of-the-art methods on the NTU-60 and NTU-120 datasets. The proposed DVTA achieves excellent performance in zero-shot learning. Compared with SMIE, DVTA achieves 1.3\% and 3.96\% performance improvement on the 55/5 split and 48/12 split of the NTU-60 dataset, respectively. For the NTU-120 dataset, DVTA achieves \textbf{9.15}\% and \textbf{6.51}\% on the 110/10 split and 96/24 split, respectively. 
“DVTA w/o SDE” means zero-shot learning on the pure network architecture without Semantic Description Enhancement. The performance of all settings exceeds the SMIE, reflecting the superiority of our network architecture.
It is noteworthy that as the number of unseen classes increases, zero-shot learning becomes more challenging. However, our DVTA still performs remarkably well, especially on the more challenging 110/10 split and 96/24 split, which fully demonstrates the significant advantage of our model in handling scenarios with more unseen classes.

\begin{table}[htbp]
	\centering
	\caption{Comparison of state-of-the-art methods with DVTA on the NTU-60 and NTU-120 datasets.}
	\begin{tabular}{l|c|c|c|c}
		\toprule
		Method & \multicolumn{2}{c|}{NTU-60} & \multicolumn{2}{c}{NTU-120} \\
		\cmidrule{2-5}    Split & 55/5  & 48/12 & 110/10 & 96/24 \\
		\midrule
		DeViSE~\cite{frome2013devise} & 60.72 & 24.51 & 47.49 & 25.74 \\
		RelationNet~\cite{jasani2019skeleton} & 40.12 & 30.06 & 52.59 & 29.06 \\
		ReViSE~\cite{hubert2017learning} & 53.91 & 17.49 & 55.04 & 32.38 \\
		JPoSE~\cite{wray2019fine} & 64.82 & 28.75 & 51.93 & 32.44 \\
		CADA-VAE~\cite{schonfeld2019generalized} & 76.84 & 28.96 & 59.53 & 35.77 \\
		SynSE~\cite{gupta2021syntactically} & 75.81 & 33.30 & 62.69 & 38.70 \\
		SMIE~\cite{zhou2023zero}  & 77.98 & 40.18 & 65.74 & 45.30 \\
		PURLS~\cite{zhu2024part}  & 79.23 & 40.99 & 71.95 & \textbf{52.01} \\
		\midrule
		DVTA w/o SDE& 78.14& 41.67& 72.30& 48.53\\
		\textbf{DVTA}& \textbf{79.28}& \textbf{44.14}& \textbf{74.89}& 51.81\\
		\bottomrule
	\end{tabular}
	\label{tab:sota}
\end{table}

\begin{table}[htbp]
		\centering
		\caption{Optimized experimental setting on NTU-60, NTU-120, and PKU-MMD datasets.}
		\begin{tabular}{l|c|c|c}
			\toprule
			Methods & NTU-60 & NTU-120 & PKU-MMD \\
			\cmidrule{2-4}    Split & 55/5  & 110/10 & 46/5 \\
			\midrule
			DeViSE~\cite{frome2013devise} & 49.80 & 44.59 & 47.94 \\
			RelationNet~\cite{jasani2019skeleton} & 48.16 & 40.55 & 51.97 \\
			ReViSE~\cite{hubert2017learning} & 56.97 & 49.32 & 65.65 \\
			SMIE~\cite{zhou2023zero}  & 63.57 & 56.37 & 67.15 \\
			\midrule
			\textbf{DVTA} & \textbf{74.03} & \textbf{60.33} & \textbf{77.06} \\
			\bottomrule
		\end{tabular}
		\label{tab:Optimized experimental setting}
	\end{table}

\noindent\textbf{Optimized Experimental Setting Results}: To better validate the effectiveness of our method in zero-shot learning, we conduct comparison experiments under an optimized setting~\cite{zhou2023zero}. Specifically, we use three large-scale skeleton datasets, i.e., NTU-60, NTU-120, and PKU-MMD as benchmark datasets. For each dataset, we conduct three randomized splits and average the results to reduce biases introduced by different class splits. The seen/unseen class splits are set as 55/5, 110/10 and 46/5, respectively, following the class split settings in SMIE~\cite{zhou2023zero}. To better focus on evaluating the effectiveness of the alignment model, we adopt the classical ST-GCN~\cite{yan2018spatial} as the visual feature extractor to minimize the impact of complex feature extractor architectures on the results. Sentence-BERT is used for semantic feature extraction and evaluation. The comparison results are shown in Table~\ref{tab:Optimized experimental setting}. Our method outperforms others across all three datasets, achieving improvements of \textbf{10.46}\%, \textbf{3.96}\%, and \textbf{9.91}\%, respectively. The results indicate that DVTA effectively aligns the skeleton and text modalities, and the dual-alignment approach is more conducive to zero-shot learning.

\begin{table*}
  \centering
  \caption{Influence of different modules on the NTU-60 and NTU-120 datasets.}
  \resizebox{\textwidth}{!}{
    \begin{tabular}{c|ccc|ccc|c|ccc|c}
    \toprule
          & \multicolumn{3}{c|}{modules} & \multicolumn{4}{c|}{NTU-60: 55/5 Split (Acc \%)} & \multicolumn{4}{c}{NTU-120: 110/10 Split (Acc \%)} \\
    \cmidrule{2-12}
          & SDE   & DA   & AA    & Split 1 & Split 2 & Split 3 & Average & Split 1 & Split 2 & Split 3 & Average \\
    \midrule
    (1) &       &       &       & 61.53 & 57.14 & 59.51 & 59.39 & 47.34 & 45.67 & 65.63 & 52..88 \\
    (2) & \checkmark &       &       & 69.24 & 64.12 & 76.86 & 70.07 & 53.98 & 41.52 & 73.27 & 56.25 \\
    (3) & \checkmark & \checkmark &       & 69.87 & 65.17 & 82.64 & 72.56 & 55.23 & 44.17 & 77.98 & 59.12 \\
    (4) & \checkmark &       & \checkmark & 75.99 & 65.23 & 73.56 & 71.59 & 55.18 & 50.88 & 72.57 & 59.31 \\
    (5) & \checkmark & \checkmark & \checkmark & 72.36 & 66.67 & 83.08 & \textbf{74.03} & 59.4  & 44.17 & 77.44 & \textbf{60.33} \\
    \bottomrule
    \end{tabular}
    }
  \label{tab:Influence of different modules}
\end{table*}

\subsection{Ablation Study and Visualizations} \label{sec:ablation}


\noindent\textbf{Influence of Modules}:
To evaluate the effectiveness of each module in the Dual Visual-Text Alignment (DVTA), we conduct a systematic ablation study. The experiments are performed under the optimized settings, with three split tests on the NTU-60 and NTU-120 datasets, and the average performance is reported, as shown in Table~\ref{tab:Influence of different modules}.

\begin{itemize}
    \item SDE (Semantic Descriptions Enhancement): In this setting, only SDE is used without the DA module for modality mapping. Instead, a simple linear layer maps the skeleton features to the dimension of the text features. The SDE facilitates mutual understanding and alignment between modalities, resulting in improved performance.
    
    \item DA (Direct Alignment): The introduction of the DA module, which utilizes a designed deep neural network to embed both modalities into the same space, significantly enhances performance. Compared to using only SDE, the performance improves by \textbf{2.49\%} and \textbf{2.87\%} on the 55/5 split of NTU-60 and the 110/10 split of NTU-120, respectively. This indicates that the deep feature projector effectively reduces the distance between modalities, facilitating the recognition of more unseen classes.
    
    \item AA (Augmented Alignment): When the AA module is introduced for modality distribution alignment, performance improves by \textbf{1.52\%} and \textbf{3.06\%} on the two datasets, respectively. While the performance on the NTU-60 dataset is slightly lower than that of the DA module, it performs better on the NTU-120 dataset. This suggests that the AA module may be more effective on larger-scale datasets.
    
    \item DVTA (Dual Visual-Text Alignment): The final DVTA model, which applies the dual alignment network, achieves the best performance with increases of \textbf{3.96\%} and \textbf{4.08\%}. The experimental results demonstrate that the dual alignment strategy with augmented supervision is consistently effective in zero-shot learning.
\end{itemize}

\begin{table}
  \small
  \centering
  \caption{Effect on different settings of Direct Alignment module.}
    \begin{tabular}{c|cc|c|c}
    \toprule
          & \multicolumn{2}{c|}{DA Components} & NTU-60(\%) & NTU-120(\%) \\
    \cmidrule{2-5}
          & Semantic & Visual & Average & Average \\
    \midrule
    (1)   &       &       & 70.07 & 56.25 \\
    (2)   & \checkmark &       & 67.35 & 56.36 \\
    (3)   &       & \checkmark & 71.6  & 58.83 \\
    (4)   & \checkmark & \checkmark & \textbf{72.56} & \textbf{59.12} \\
    \bottomrule
    \end{tabular}
  \label{tab:JEL}
\end{table}

\begin{figure}[htbp]
    \centering
    \includegraphics[width=1\linewidth]{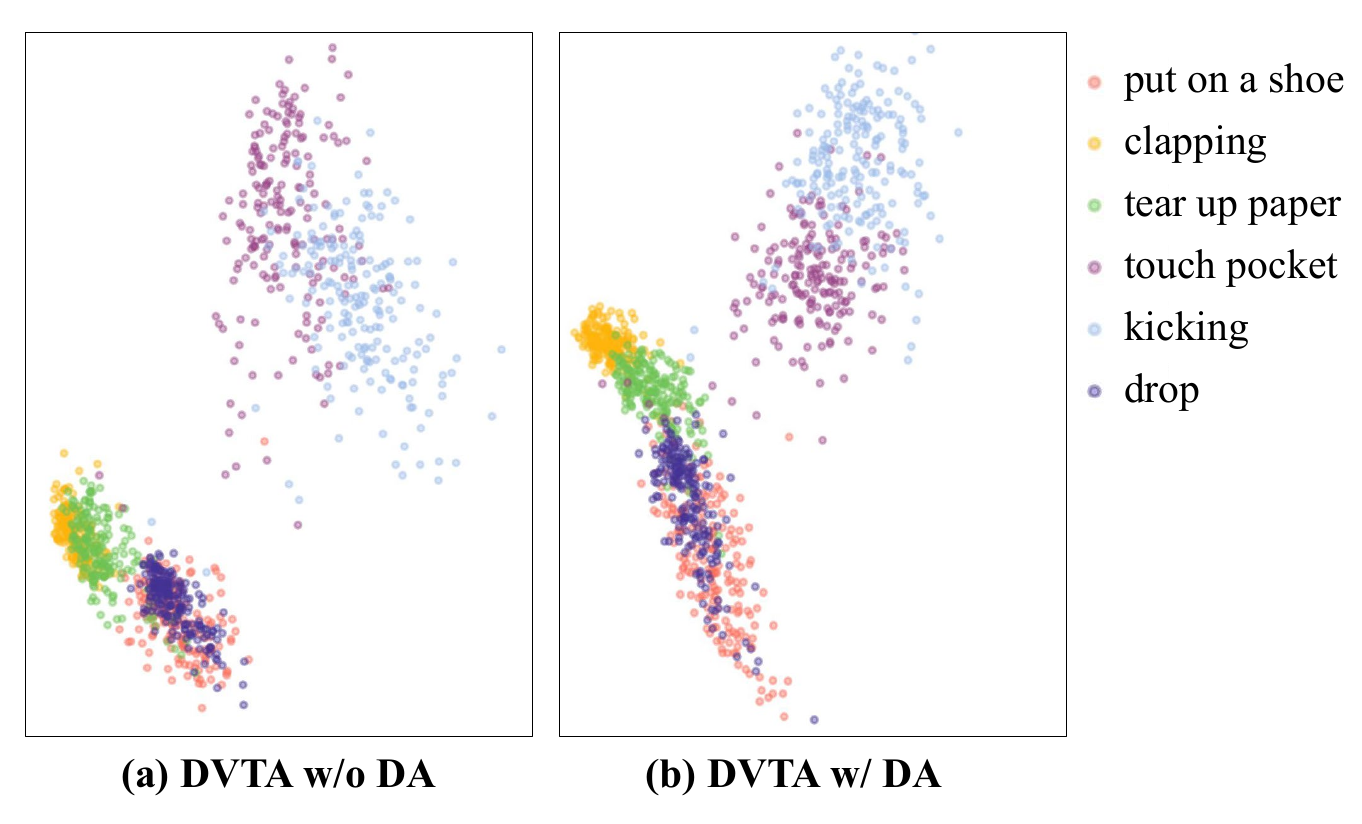}
    \caption{The PCA visualization of latent representations without and with the DA module. The six action categories are randomly selected from unseen classes in the 48/12 split of the NTU-60 dataset.}
    \label{fig:JEL_effect}
\end{figure}


\noindent\textbf{Effect of Direct Alignment module}: 
The Direct Alignment (DA) module maps visual and text features into an embedding space, helping to reduce the gap between the two modalities and thus improving the recognition accuracy of unseen classes. As shown in Figure \ref{fig:JEL_effect}, we use Principal Component Analysis (PCA) to compare skeleton feature representations without and with the DA module. Compared to not using the DA module, the skeleton features exhibit a more distinct distribution among different action categories when the DA module is applied. 

The DA module can be applied as an independent model for zero-shot skeleton action recognition, optimized through training with the $L_1$ loss function. This module comprises a deep nonlinear mapping projector for skeleton features and a linear layer for text features, allowing for the investigation of the utility of these components through ablation experiments. The results of the ablated experiments on the DA module's components are shown in Table \ref{tab:JEL}, where ``Semantic'' denotes the use of the text mapping, and ``Visual'' denotes the use of the designed deep nonlinear mapping projector for skeleton features. The results indicate that using the complete DA module more effectively reduces the gap between modalities, thereby improving alignment quality.

\begin{table}
  \centering
  \caption{Effect on different settings of the hyperparameter $\gamma$ on Augmented Alignment module.}
    \begin{tabular}{l|cc|cc}
    \toprule
    \multirow{2}[0]{*}{\textbf{$\gamma$} on AA} & \multicolumn{2}{c|}{NTU-60(\%)} & \multicolumn{2}{c}{NTU-120(\%)} \\
    \cmidrule{2-5}
          & 55/5  & 48/12 & 110/10 & 96/24 \\
    \midrule
    None  & 77.65 & 39.18 & 72.08 & 44.20 \\
    Sigmoid & 77.93 & 39.31 & 71.88 & 45.52 \\
    0.005 & 78.07 & 40.67 & 72.27 & 47.76 \\
    0.01  & \textbf{78.24} & \textbf{41.67} & 72.30 & \textbf{48.53} \\
    0.1   & 78.21 & 40.92 & 71.90 & 45.80 \\
    0.5   & 77.93 & 39.91 & \textbf{73.48} & 45.11 \\
    \bottomrule
    \end{tabular}%
  \label{tab:leakiness}%
\end{table}%

\begin{figure}
    \centering
    \includegraphics[width=1\linewidth]{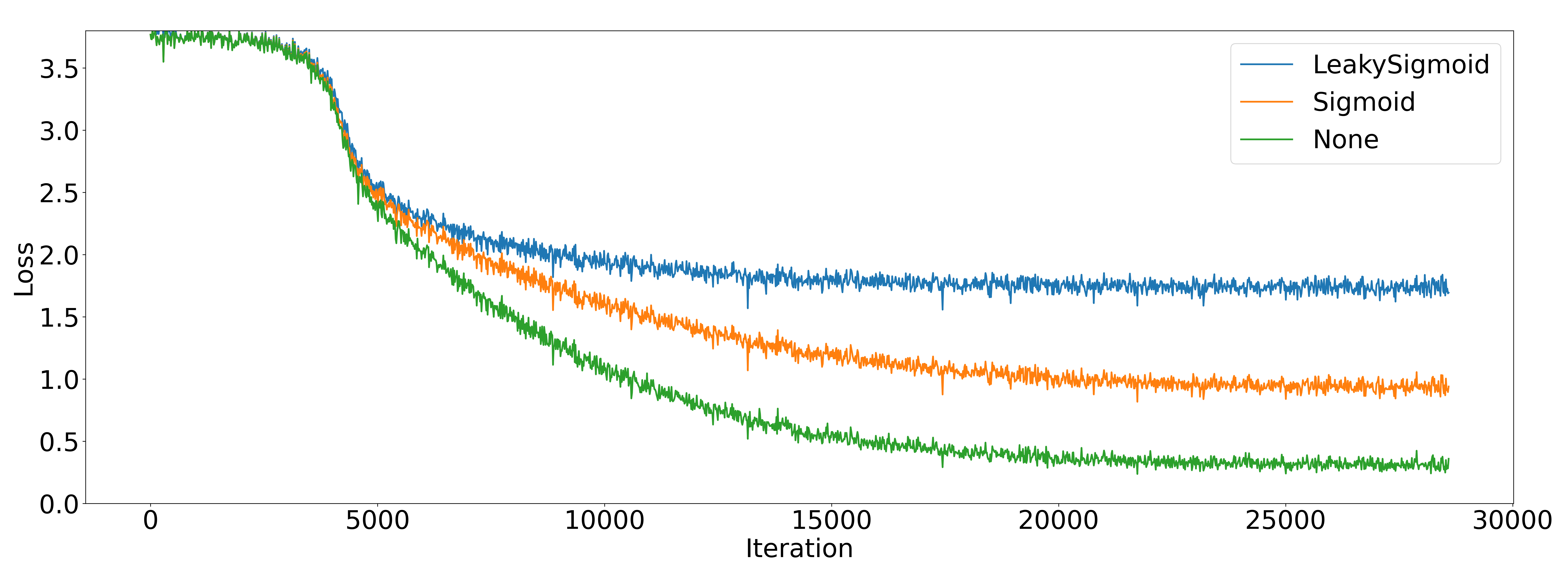}
    \caption{Convergence curves of LeakySigmoid, Sigmoid and None losses. They are trained on the 55/5 split of NTU-60 dataset, where the LeakySigmoid function converges first.}
    \label{fig:loss}
\end{figure}

\noindent\textbf{Effect of Augmented Alignment module}: In the Augmented Alignment (AA) module, we design a LeakySigmoid function to convert the output value of the network $E$ into a similarity score $G$. It has a hyperparameter, $\gamma$, which controls the slope of the function at negative values. To determine the optimal choice of $\gamma$ and demonstrate the effectiveness of the LeakySigmoid function, we conducted experiments on all four splits based on the complete model. The experimental results are shown in Table \ref{tab:leakiness}, where ``None'' indicates no utilization of any transformation function, and ``Sigmoid'' signifies the application of the Sigmoid function to output similarity scores. Through comparison, it is found that the performance of the LeakySigmoid function consistently surpasses that of the Sigmoid function, regardless of the value set for the $\gamma$ hyperparameter. This discovery effectively validates the usefulness of the LeakySigmoid function in the AA module. Specifically, when $\gamma$ was set to 0.01, the model exhibits the best performance on two splits. To ensure consistency and comparability in our experiments, we uniformly set the $\gamma$ hyperparameter to 0.01 throughout the entire experimental process.

We compare the effects of three different functions on the model on the 55/5 split: using the LeakySigmoid function, the Sigmoid function, and no function (None), resulting in accuracies of 78.14\%, 77.93\%, and 77.65\% respectively. In Figure \ref{fig:loss}, we show the loss curves for the three methods. It can be observed that the LeakySigmoid function not only provides the best performance but also accelerates the convergence of the network, validating its effectiveness.
\begin{table}
  \centering
  \caption{Ablation on Semantic Descriptions Enhancement module.}
    \begin{tabular}{cc|cc|cc}
    \toprule
    \multirow{2}[0]{*}{Encoder} & \multirow{2}[0]{*}{SDE} & \multicolumn{2}{c|}{NTU-60(\%)} & \multicolumn{2}{c}{NTU-120(\%)} \\
    \cmidrule{3-6}
          &       & 55/5  & 48/12 & 110/10 & 96/24 \\
    \midrule
    \multirow{2}[0]{*}{CLIP} & w/o   & 78.50 & 39.20 & 61.27 & 46.77 \\
          & w     & \textbf{79.64} & \textbf{47.48} & 64.11 & 48.18 \\
    \midrule
    \multirow{2}[0]{*}{BERT} & w/o   & 78.14 & 41.67 & 72.30 & 48.53 \\
          & w     & 79.28 & 44.14 & \textbf{74.89} & \textbf{51.81} \\
    \bottomrule
    \end{tabular}%
  \label{tab:SDE}%
\end{table}%

\begin{figure}[htbp]
    \centering
    \includegraphics[width=1\linewidth]{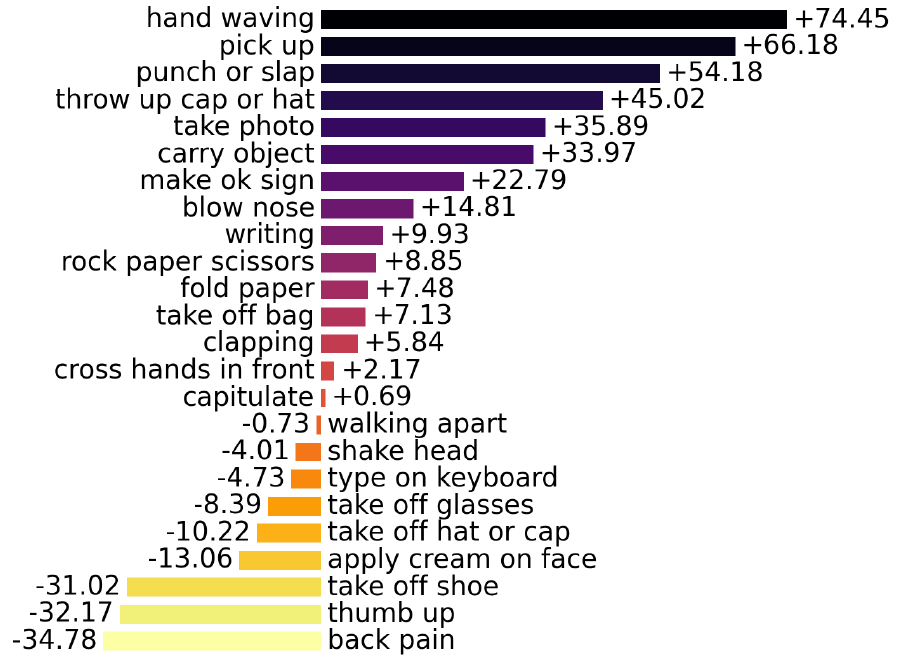}
    \caption{Accuracy change chart of action categories on the 96/24 split of the NTU-120 dataset using the Semantic Description Enhancement module.}
    \label{fig:bar}
\end{figure}

\noindent\textbf{Effect of Semantic Descriptions Enhancement module}: 
We encode the action text using different text encoders. Table~\ref{tab:SDE} presents the SDE ablation study results using Sentence-BERT and CLIP as semantic feature extractors on the NTU-60 and NTU-120 datasets. The results indicate that the use of the Semantics Description Enhancement (SDE) module leads to significant performance improvements across all split tests, regardless of the encoder used. For the NTU-120 dataset, under the 96/24 split, the accuracy of action labels and the accuracy when enhanced with action description encoding are 48.53\% and 51.81\%, respectively, using the BERT encoder. Figure \ref{fig:bar} shows the per-class accuracy changes after applying the SDE module. The visualization reveals substantial improvements in the accuracy of certain hand-related action categories, such as ``waving'', ``picking up'', and ``punching and kicking''.
\begin{table}
  \small
  \centering
  \caption{Performance of different skeleton and text encoders on the NTU-60 and NTU-120 datasets.}
   \resizebox{0.49\textwidth}{!}{
    \begin{tabular}{cc|cc|cc}
    \toprule
    \multicolumn{1}{c}{\multirow{2}[4]{*}{\makecell[c]{Skeleton\\Encoder}}} & \multicolumn{1}{c|}{\multirow{2}[4]{*}{\makecell[c]{Text\\Encoder}}} & \multicolumn{2}{c|}{NTU-60(\%)} & \multicolumn{2}{c}{NTU-120(\%)} \\
\cmidrule{3-6}          &       & 55/5  & 48/12 & 110/10 & 96/24 \\
    \midrule
    ST-GCN~\cite{yan2018spatial} & BERT  & 73.99 & 41.32 & 63.22 & 47.99 \\
    ST-GCN~\cite{yan2018spatial} & CLIP  & 75.51 & 43.37 & 61.91 & 44.54 \\
    Shift-GCN~\cite{cheng2020skeleton} & BERT  & 79.28 & 44.14 & 74.89 & 51.81 \\
    Shift-GCN~\cite{cheng2020skeleton} & CLIP  & 79.64 & 47.48 & 64.11 & 48.18 \\
    CTR-GCN~\cite{chen2021channel} & BERT  & 81.07 & 45.92 & \textbf{76.86} & 54.70 \\
    CTR-GCN~\cite{chen2021channel} & CLIP  & \textbf{82.55} & \textbf{48.40} & 68.26 & \textbf{56.97} \\
    \bottomrule
    \end{tabular}}
  \label{tab:Different Encoders}%
\end{table}%

\noindent\textbf{Effect of Different Encoders}:
Table~\ref{tab:Different Encoders} reports the performance of our proposed method on the NTU-60 and NTU-120 datasets using different visual and text encoders. We adopt several widely used skeleton-based backbones as visual encoders, including ST-GCN, Shift-GCN and CTR-GCN. Sentence-BERT and CLIP are used as text encoders for semantic feature extraction. The results show that performance varies across different text encoders, possibly due to the diverse knowledge embedded in them, some of which may not directly relate to action class semantics. Nevertheless, irrespective of the text encoder used, our method consistently achieves better performance when equipped with more advanced visual feature extractors. This suggests that the proposed approach exhibits stronger zero-shot learning capabilities when combined with more sophisticated skeleton backbones.

\begin{table}
  \small
  \centering
  \caption{Experiments with the loss function on the NTU-60 dataset with different temperature parameters.}
    \begin{tabular}{c|c|c|c|c}
    \toprule
    \multirow{2}[0]{*}{Loss function} & \multicolumn{4}{c}{NTU-60(\%)} \\
    \cmidrule{2-5}
          & $\tau$ = 0.07 & $\tau$ = 0.1 & $\tau$ = 0.5 & $\tau$ = 1 \\
    \midrule
    InfoNCE & 71.5  & 73.04 & 72.77 & 70.08 \\
    SoftmaxCE & 71.5  & 73.09 & 72.87 & 70.09 \\
    KLD   & 72.67 & \textbf{74.03} & 73.91 & 71.14 \\
    \bottomrule
    \end{tabular}%
  \label{tab:Loss function}%
\end{table}%

\noindent\textbf{Effect of the Loss Functions}:
We evaluate the InfoNCE, SoftmaxCE, and KLD loss functions under different temperature parameters, with the experimental results shown in Table~\ref{tab:Loss function}. The performance of these losses is compared based on the average results across the three splits in the NTU-60 dataset. The KLD loss demonstrates the best performance, confirming the effectiveness of using more positive samples for contrastive learning. In contrast, the InfoNCE and SoftmaxCE losses define the positive samples as only the corresponding semantic feature for each visual feature, resulting in fewer positive samples and thus suboptimal performance.
\begin{table}[tbp]
  \small
  \centering
  \caption{Computational efficiency comparison. Params, FLOPs, and FPS represent the number of learnable parameters, the floating point operations per sample, and the number of samples processed per second during inference, respectively.}
    \begin{tabular}{c|ccc|c}
    \toprule
    Method & Params & FLOPs & FPS   & NTU-60 \\
    \midrule
    SMIE~\cite{zhou2023zero}  & 1.57M & 3.14M & 1623  & 40.18 \\
    PURLS~\cite{zhu2024part} & 7.58M & 17.32M & 497   & 40.99 \\
    DVTA  & 4.88M & 4.87M & 1335  & 44.14 \\
    \bottomrule
    \end{tabular}%
  \label{tab:Computational efficiency}%
\end{table}%

\noindent\textbf{Computational Efficiency Analysis}:
As shown in Table~\ref{tab:Computational efficiency}, we compare the Params, FLOPs, and FPS of our method (DVTA) with SMIE and PURLS, evaluated on the same hardware setup (NVIDIA GeForce RTX 4090). For these metrics, we measure the computational efficiency of the alignment model independently, using pre-extracted visual and text features as input. Inference time is measured on the NTU-60 dataset of 110/10 split by running the models on the same number of test samples, and FPS is computed by dividing the total inference time by the number of samples. The results indicate that both SMIE and DVTA maintain high computational efficiency. PURLS exhibits lower efficiency due to its extensive use of attention mechanisms. Overall, DVTA achieves competitive performance while maintaining relatively low computational complexity.


\begin{figure}[htbp]
    \centering
    \includegraphics[width=\linewidth]{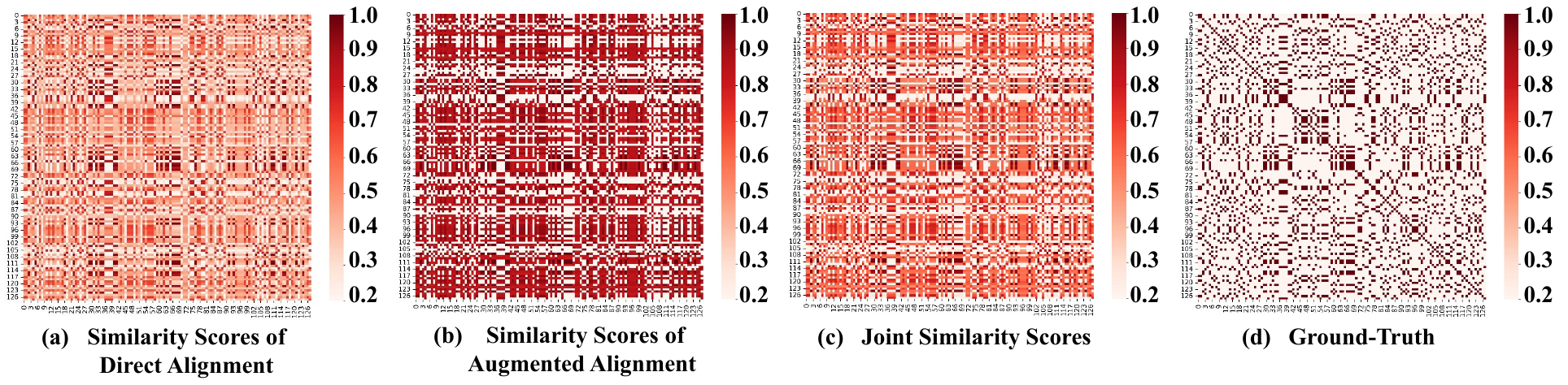}
    \caption{Visualization of the similarity score matrices for each module and the ground-truth on the 55/5 split of the NTU-60 dataset.}
    \label{fig:matrix}
\end{figure}

\noindent\textbf{Visualization of the Similarity Score}: In Figure \ref{fig:matrix}, we visualize the similarity scores for each module as well as the ground-truth. It can be observed that the ground-truth in a batch of 128 is not a standard diagonal matrix and each sample has multiple corresponding labels forming positive samples. To this end, KL divergence is more suitable as the loss function compared to cross-entropy which is typically used for 1-in-N classification problems. The similarity score matrix of Direct Alignment module is already quite close to the ground-truth contours, indicating promising potential for zero-shot learning of unseen classes. The Augmented Alignment module amplifies the score differences between positive and negative samples, significantly enhancing the model’s discriminative ability. When the similarity score matrices of the two modules are combined, it enhances the robustness of the model and achieves better results.
\begin{figure}[htbp]
    \centering
    \includegraphics[width={1\linewidth}]{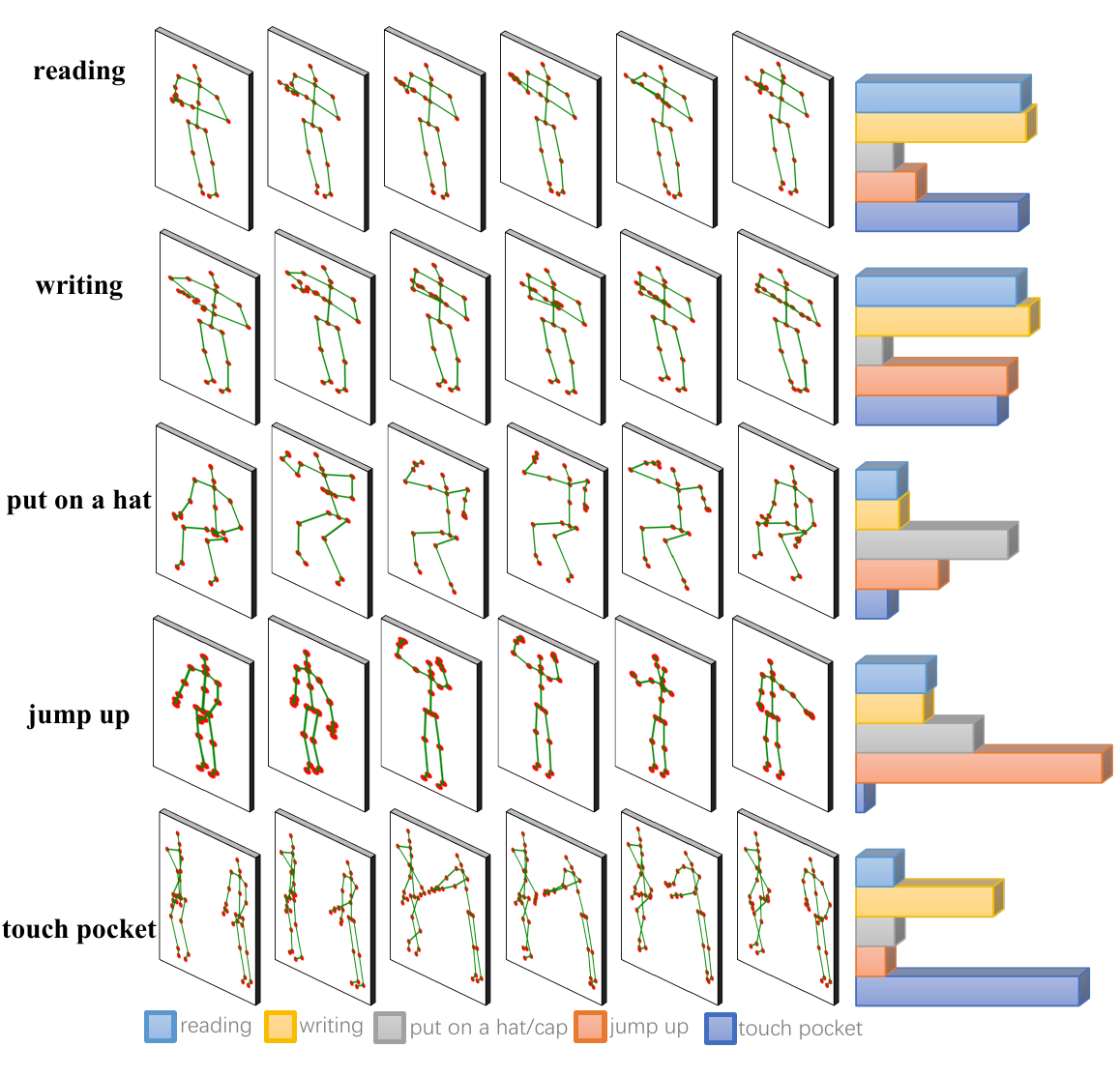}
    \caption{The Zero-shot Recognition Results on the 55/5 split of the NTU-60 dataset.}
    \label{fig:skeleton}
\end{figure}

\noindent\textbf{Qualitative Results and Analysis}: Figure \ref{fig:skeleton} illustrates the skeleton visualization of the 55/5 split on the NTU-60 dataset. The left side showcases the skeleton dynamics of five unseen class actions, while the right side displays the action prediction scores. The action "put on a hat", which involves raising the hands above the head, is similar to "jump up", however, our model makes the correct prediction. Furthermore, "jump up" ranking as the second highest prediction score also indicates that our model has reasonable predictions for similar actions. The prediction scores for other actions also support the effectiveness of the method. 
\section{Conclusion}
This work introduces a novel Dual Visual-Text Alignment (DVTA) for skeleton-based zero-shot action recognition. DVTA mainly comprises two alignment modules, Direct Alignment (DA) module and Augmented Alignment (AA) module, and the Semantic Description Enhancement (SDE) submodule. Our approach considerably outperforms existing zero-shot state-of-the-art methods on NTU RGB+D 60, NTU RGB+D 120 and PKU-MMD. Detailed ablation studies demonstrate the effectiveness of the three components as well as the validity of designed projectors and the LeakySigmoid activation function. Visualizations of similarity scores between skeleton features and text features confirm the complementarity between the DA and the Augmented Alignment. We believe that the proposed method can be applied to the general task of zero-shot or open-set action recognition. However, a limitation of our approach is its reliance on pre-extracted features. In the future, we will focus on learning joint visual-text representations end-to-end, allowing for more dynamic feature extraction and improving the applicability of the method to real-time and large-scale scenarios.

\section*{Acknowledgement}
This research was supported by the Jiangsu Science Foundation (BK20230833), the National Science Foundation of China (62302093, 62172090), Start-up Research Fund of Southeast University under Grant RF1028623097, and the Big Data Computing Center of Southeast University.

\bibliographystyle{elsarticle-num}
\bibliography{ref.bib}

\begin{thebibliography}{10}
\expandafter\ifx\csname url\endcsname\relax
  \def\url#1{\texttt{#1}}\fi
\expandafter\ifx\csname urlprefix\endcsname\relax\def\urlprefix{URL }\fi
\expandafter\ifx\csname href\endcsname\relax
  \def\href#1#2{#2} \def\path#1{#1}\fi

\bibitem{xu2017transductive}
X.~Xu, T.~Hospedales, S.~Gong, Transductive zero-shot action recognition by
  word-vector embedding, International Journal of Computer Vision 123 (2017)
  309--333.

\bibitem{xu2015semantic}
X.~Xu, T.~Hospedales, S.~Gong, Semantic embedding space for zero-shot action
  recognition, in: IEEE International Conference on Image Processing (ICIP),
  IEEE, 2015, pp. 63--67.

\bibitem{wang2017zero}
Q.~Wang, K.~Chen, Zero-shot visual recognition via bidirectional latent
  embedding, International Journal of Computer Vision 124 (2017) 356--383.

\bibitem{wang2017modeling}
H.~Wang, L.~Wang, Modeling temporal dynamics and spatial configurations of
  actions using two-stream recurrent neural networks, in: Proceedings of the
  IEEE Conference on Computer Vision and Pattern Recognition, 2017, pp.
  499--508.

\bibitem{weng2025usdrl}
W.~Weng, H.~Wang, J.~Wang, L.~He, G.-S. Xie, Usdrl: Unified skeleton-based
  dense representation learning with multi-grained feature decorrelation, in:
  Proceedings of the AAAI Conference on Artificial Intelligence, Vol.~39, 2025,
  pp. 8332--8340.

\bibitem{zhou2023zero}
Y.~Zhou, W.~Qiang, A.~Rao, N.~Lin, B.~Su, J.~Wang, Zero-shot skeleton-based
  action recognition via mutual information estimation and maximization, in:
  Proceedings of the ACM International Conference on Multimedia, 2023, pp.
  5302--5310.

\bibitem{han2020learning}
Z.~Han, Z.~Fu, J.~Yang, Learning the redundancy-free features for generalized
  zero-shot object recognition, in: Proceedings of the IEEE/CVF Conference on
  Computer Vision and Pattern Recognition, 2020, pp. 12865--12874.

\bibitem{brattoli2020rethinking}
B.~Brattoli, J.~Tighe, F.~Zhdanov, P.~Perona, K.~Chalupka, Rethinking zero-shot
  video classification: End-to-end training for realistic applications, in:
  Proceedings of the IEEE/CVF Conference on Computer Vision and Pattern
  Recognition, 2020, pp. 4613--4623.

\bibitem{gao2023learning}
J.~Gao, Y.~Hou, Z.~Guo, H.~Zheng, Learning spatio-temporal semantics and
  cluster relation for zero-shot action recognition, IEEE Transactions on
  Circuits and Systems for Video Technology (2023).

\bibitem{qi2023energy}
C.~Qi, Z.~Feng, M.~Xing, Y.~Su, J.~Zheng, Y.~Zhang, Energy-based temporal
  summarized attentive network for zero-shot action recognition, IEEE
  Transactions on Multimedia (2023).

\bibitem{lin2023match}
W.~Lin, L.~Karlinsky, N.~Shvetsova, H.~Possegger, M.~Kozinski, R.~Panda,
  R.~Feris, H.~Kuehne, H.~Bischof, Match, expand and improve: Unsupervised
  finetuning for zero-shot action recognition with language knowledge, in:
  Proceedings of the IEEE/CVF International Conference on Computer Vision,
  2023, pp. 2851--2862.

\bibitem{wang2023deconfounding}
J.~Wang, Y.~Jiang, Y.~Long, X.~Sun, M.~Pagnucco, Y.~Song, Deconfounding causal
  inference for zero-shot action recognition, IEEE Transactions on Multimedia
  (2023).

\bibitem{yan2018spatial}
S.~Yan, Y.~Xiong, D.~Lin, Spatial temporal graph convolutional networks for
  skeleton-based action recognition, in: Proceedings of the Thirty-Second AAAI
  Conference on Artificial Intelligence and Thirtieth Innovative Applications
  of Artificial Intelligence Conference and Eighth AAAI Symposium on
  Educational Advances in Artificial Intelligence, 2018, pp. 7444--7452.

\bibitem{dai2023global}
M.~Dai, Z.~Sun, T.~Wang, J.~Feng, K.~Jia, Global spatio-temporal synergistic
  topology learning for skeleton-based action recognition, Pattern Recognition
  140 (2023) 109540.

\bibitem{zhao2024sharing}
Y.~Zhao, Q.~Gao, Z.~Ju, J.~Zhou, Y.~Guo, Sharing-net: Lightweight feedforward
  network for skeleton-based action recognition based on information sharing
  mechanism, Pattern Recognition 146 (2024) 110050.

\bibitem{yin2024spatiotemporal}
X.~Yin, J.~Zhong, D.~Lian, W.~Cao, Spatiotemporal progressive inward-outward
  aggregation network for skeleton-based action recognition, Pattern
  Recognition 150 (2024) 110262.

\bibitem{li2019actional}
M.~Li, S.~Chen, X.~Chen, Y.~Zhang, Y.~Wang, Q.~Tian, Actional-structural graph
  convolutional networks for skeleton-based action recognition, in: Proceedings
  of the IEEE/CVF Conference on Computer Vision and Pattern Recognition, 2019,
  pp. 3595--3603.

\bibitem{shi2019two}
L.~Shi, Y.~Zhang, J.~Cheng, H.~Lu, Two-stream adaptive graph convolutional
  networks for skeleton-based action recognition, in: Proceedings of the
  IEEE/CVF Conference on Computer Vision and Pattern Recognition, 2019, pp.
  12026--12035.

\bibitem{cheng2020skeleton}
K.~Cheng, Y.~Zhang, X.~He, W.~Chen, J.~Cheng, H.~Lu, Skeleton-based action
  recognition with shift graph convolutional network, in: Proceedings of the
  IEEE/CVF Conference on Computer Vision and Pattern Recognition, 2020, pp.
  183--192.

\bibitem{cheng2021extremely}
K.~Cheng, Y.~Zhang, X.~He, J.~Cheng, H.~Lu, Extremely lightweight
  skeleton-based action recognition with shiftgcn++, IEEE Transactions on Image
  Processing 30 (2021) 7333--7348.

\bibitem{chen2021channel}
Y.~Chen, Z.~Zhang, C.~Yuan, B.~Li, Y.~Deng, W.~Hu, Channel-wise topology
  refinement graph convolution for skeleton-based action recognition, in:
  Proceedings of the IEEE/CVF International Conference on Computer Vision,
  2021, pp. 13359--13368.

\bibitem{chi2022infogcn}
H.-g. Chi, M.~H. Ha, S.~Chi, S.~W. Lee, Q.~Huang, K.~Ramani, Infogcn:
  Representation learning for human skeleton-based action recognition, in:
  Proceedings of the IEEE/CVF Conference on Computer Vision and Pattern
  Recognition, 2022, pp. 20186--20196.

\bibitem{xiang2023generative}
W.~Xiang, C.~Li, Y.~Zhou, B.~Wang, L.~Zhang, Generative action description
  prompts for skeleton-based action recognition, in: Proceedings of the
  IEEE/CVF International Conference on Computer Vision, 2023, pp. 10276--10285.

\bibitem{qiu2024multi}
H.~Qiu, B.~Hou, Multi-grained clip focus for skeleton-based action recognition,
  Pattern Recognition 148 (2024) 110188.

\bibitem{wu2023spatiotemporal}
L.~Wu, C.~Zhang, Y.~Zou, Spatiotemporal focus for skeleton-based action
  recognition, Pattern Recognition 136 (2023) 109231.

\bibitem{frome2013devise}
A.~Frome, G.~S. Corrado, J.~Shlens, S.~Bengio, J.~Dean, M.~Ranzato, T.~Mikolov,
  Devise: A deep visual-semantic embedding model, Advances in Neural
  Information Processing Systems 26 (2013).

\bibitem{hubert2017learning}
Y.-H. Hubert~Tsai, L.-K. Huang, R.~Salakhutdinov, Learning robust
  visual-semantic embeddings, in: Proceedings of the IEEE International
  Conference on Computer Vision, 2017, pp. 3571--3580.

\bibitem{jasani2019skeleton}
B.~Jasani, A.~Mazagonwalla, Skeleton based zero shot action recognition in
  joint pose-language semantic space, arXiv preprint arXiv:1911.11344 (2019).

\bibitem{wray2019fine}
M.~Wray, D.~Larlus, G.~Csurka, D.~Damen, Fine-grained action retrieval through
  multiple parts-of-speech embeddings, in: Proceedings of the IEEE/CVF
  International Conference on Computer Vision, 2019, pp. 450--459.

\bibitem{schonfeld2019generalized}
E.~Schonfeld, S.~Ebrahimi, S.~Sinha, T.~Darrell, Z.~Akata, Generalized zero-and
  few-shot learning via aligned variational autoencoders, in: Proceedings of
  the IEEE/CVF Conference on Computer Vision and Pattern Recognition, 2019, pp.
  8247--8255.

\bibitem{gupta2021syntactically}
P.~Gupta, D.~Sharma, R.~K. Sarvadevabhatla, Syntactically guided generative
  embeddings for zero-shot skeleton action recognition, in: IEEE International
  Conference on Image Processing (ICIP), IEEE, 2021, pp. 439--443.

\bibitem{zhu2024part}
A.~Zhu, Q.~Ke, M.~Gong, J.~Bailey, Part-aware unified representation of
  language and skeleton for zero-shot action recognition, in: Proceedings of
  the IEEE/CVF Conference on Computer Vision and Pattern Recognition, 2024, pp.
  18761--18770.

\bibitem{chen2024fine}
Y.~Chen, J.~Guo, T.~He, X.~Lu, L.~Wang, Fine-grained side information guided
  dual-prompts for zero-shot skeleton action recognition, in: Proceedings of
  the 32nd ACM International Conference on Multimedia, 2024, pp. 778--786.

\bibitem{xu2025information}
H.~Xu, Y.~Gao, J.~Li, X.~Gao, An information compensation framework for
  zero-shot skeleton-based action recognition, IEEE Transactions on Multimedia
  (2025).

\bibitem{brown2020language}
T.~Brown, B.~Mann, N.~Ryder, M.~Subbiah, J.~D. Kaplan, P.~Dhariwal,
  A.~Neelakantan, P.~Shyam, G.~Sastry, A.~Askell, et~al., Language models are
  few-shot learners, Advances in Neural Information Processing Systems 33
  (2020) 1877--1901.

\bibitem{wang2021actionclip}
M.~Wang, J.~Xing, Y.~Liu, Actionclip: A new paradigm for video action
  recognition (2021).
\newblock \href {http://arxiv.org/abs/2109.08472} {\path{arXiv:2109.08472}}.

\bibitem{shahroudy2016ntu}
A.~Shahroudy, J.~Liu, T.-T. Ng, G.~Wang, Ntu rgb+ d: A large scale dataset for
  3d human activity analysis, in: Proceedings of the IEEE Conference on
  Computer Vision and Pattern Recognition, 2016, pp. 1010--1019.

\bibitem{liu2019ntu}
J.~Liu, A.~Shahroudy, M.~Perez, G.~Wang, L.-Y. Duan, A.~C. Kot, Ntu rgb+ d 120:
  A large-scale benchmark for 3d human activity understanding, IEEE
  Transactions on Pattern Analysis and Machine Intelligence 42~(10) (2019)
  2684--2701.

\bibitem{liu2017pku}
C.~Liu, Y.~Hu, Y.~Li, S.~Song, J.~Liu, Pku-mmd: A large scale benchmark for
  skeleton-based human action understanding, in: Proceedings of the Workshop on
  Visual Analysis in Smart and Connected Communities, Association for Computing
  Machinery, 2017, p. 1–8.

\bibitem{li20213d}
L.~Li, M.~Wang, B.~Ni, H.~Wang, J.~Yang, W.~Zhang, 3d human action
  representation learning via cross-view consistency pursuit, in: Proceedings
  of the IEEE/CVF Conference on Computer Vision and Pattern Recognition, 2021,
  pp. 4741--4750.

\end{thebibliography}

\end{document}